\documentclass[11pt,a4paper]{article}
\usepackage{acl2024} 
\usepackage{times}
\usepackage{latexsym}

\usepackage{amsmath}
\usepackage{amsfonts}
\usepackage[linesnumbered,ruled,vlined]{algorithm2e}
\usepackage{graphicx} 
\usepackage{booktabs}
\usepackage{array}
\usepackage{enumitem}
\usepackage{amsthm}
\usepackage{algpseudocode}
\usepackage[switch]{lineno}
\usepackage[utf8]{inputenc}
\usepackage{eqparbox}
\usepackage[nopar]{lipsum}
\usepackage{multirow}
\usepackage{makecell}
\usepackage{xcolor}
\usepackage{hhline} 
\usepackage{microtype}
\usepackage{diagbox}

\usepackage{adjustbox}

\usepackage{relsize}

\usepackage{booktabs,multirow,array}
\newcolumntype{N}{@{}m{0pt}@{}}

\usepackage[many]{tcolorbox}
\newtcolorbox{fancyquotes}{%
    enhanced jigsaw, 
    breakable,      
    frame hidden,   
    left=0.5cm,       
    right=0.1cm,      
    overlay={%
        \node [scale=8,
            text=black,
            inner sep=0pt,] at ([xshift=-1cm,yshift=-1cm]frame.north west){}; 
        \node [scale=8,
            text=black,
            inner sep=0pt,] at ([xshift=1cm]frame.south east){};  
            },
                parbox=false,
}

\usepackage{mathtools, nccmath}
\usepackage{scrextend}
\deffootnote[.25in]{.25in}{.15in}{\makebox[.25in][r]{\thefootnotemark .\hspace{.15in}}}

\makeatletter

\newtheorem*{proof*}{Proof}

\usepackage{listings}
\usepackage{color}
\definecolor{codegreen}{rgb}{0.3,0.5,0.0}
\def\@fnsymbol#1{\ensuremath{\ifcase#1\or \dagger\or *\or \ddagger\or
   \mathsection\or \mathparagraph\or \|\or **\or \dagger\dagger
   \or \ddagger\ddagger \else\@ctrerr\fi}}
   
\renewcommand{\arraystretch}{1.5}
\newcolumntype{C}[1]{>{\centering\let\newline\\\arraybackslash\hspace{0pt}}m{#1}}
\newcommand\ChangeRT[1]{\noalign{\hrule height #1}}

\newcommand\Vtextvisiblespace[1][.3em]{%
  \mbox{\kern.06em\vrule height.3ex}%
  \vbox{\hrule width#1}%
  \hbox{\vrule height.3ex}}

\ExplSyntaxOn
\NewExpandableDocumentCommand { \ValuePlusOne } { m } 
  { \int_eval:n { \int_use:c { c @ #1 } + 1 } }
\NewExpandableDocumentCommand { \Sec } { m } 
  { \fp_eval:n { secd ( #1 ) } }
\NewDocumentCommand { \Rot } { m }
  { 
    \hbox_to_wd:nn { 1 em }
      { 
        \hbox_overlap_right:n 
          { 
            \skip_horizontal:n { \fp_to_dim:n { 7 * cosd (\Angle) } } 
            \rotatebox{\Angle}{#1}
          } 
      } 
  }
\ExplSyntaxOff

\def\Angle{45}
    
\bigskip
\def\Angle{90}

\title{Extensible Embedding: A Flexible Multipler For LLM's Context Length}

\author{Ninglu Shao$^{1,2}$\thanks{Co-first and co-corresponding author.}, ~ Shitao Xiao$^1$\footnotemark[1], ~ Zheng Liu$^{1}$\footnotemark[1], ~ Peitian Zhang$^{1,2}$ \\
  1: Beijing Academy of Artificial Intelligence \\
  2: Gaoling School of Artificial Intelligence, Renmin University \\
  \texttt{rainym00d@163.com} \ \ \  \texttt{stxiao@baai.ac.cn} \ \ \  \texttt{zhengliu1026@gmail.com} 
}

\begin{document}
\maketitle 

\begin{abstract}
Large language models (LLMs) call for extension of context to handle many critical applications. However, the existing approaches are prone to expensive costs and inferior quality of context extension. In this work, we propose \textbf{Extensible Embedding}\footnote{https://github.com/FlagOpen/FlagEmbedding}, which realizes high-quality extension of LLM's context with strong flexibility and cost-effectiveness. Extensible embedding stand as an enhancement of typical token embedding, which represents the information for an extensible scope of context instead of a single token. By leveraging such compact input units of higher information density, the LLM can access to a vast scope of context even with a small context window. Extensible embedding is systematically optimized in architecture and training method, which leads to multiple advantages. 1) High flexibility of context extension, which flexibly supports ad-hoc extension of diverse context lengths. 2) Strong sample efficiency of training, which enables the embedding model to be learned in a cost-effective way. 3) Superior compatibility with the existing LLMs, where the extensible embedding can be seamlessly introduced as a plug-in component. Comprehensive evaluations on long-context language modeling and understanding tasks verify extensible embedding as an effective, efficient, flexible, and compatible method to extend the LLM's context. 

\end{abstract}

\section{Introduction}
Large language models (LLMs) need to process long-sequence data in order to accomplish many critical tasks, like RAG and long-doc reading comprehension. Unfortunately, the existing LLMs are limited by their context windows, which are far from enough to fully cover the input data in corresponding scenarios. To overcome this limitation, people resort to fine-tuning to extend the LLM's context window \cite{chen2023longlora,longchat2023,peng2023yarn}. Despite the popularity in practice, the fine-tuning approaches will lead to huge training and inference costs. Besides, the fine-tuning over long-sequence data is likely to impair the LLM's original capability on shorter contexts, which is unfavorable to the practical usage. Although there are other alternative ways to establish long contexts, e.g., sparse attention~\cite{child2019sparse_transformers,beltagy2020longformer,zaheer2020bigbird}, stream processing~\cite{xiao2023streamingllm,han2023lm_infinite}, retrieval \cite{xu2023retrieval_meets,wu2022memorizing,tworkowski2023focused}, the existing solutions are prone to problems, like inferior extension quality or incompatibility with the existing LLMs. 

It's usually believed that the size of context window, e.g., 4096 for LLaMA-2 \cite{touvron2023llama-b}, is equivalent to the maximum of tokens the LLM can perceive. However, we challenge this common belief by arguing that the size of context window is just a constraint of the input units instead of the limit of context the LLM can perceive. Based on this argument, we propose a new method called \textbf{Extensible Embedding} to facilitate the utilization of long context for LLMs. It stands as an enhancement of typical token embeddings, which is used to represent the information for an extensible scope of context, e.g., multiple words or a sentence. Therefore, it can possess a much higher information density than token embeddings. On top of such compact form of representations, the LLM will be able to access to the information from a vast context with its original context window. 

The extensible embedding is realized by a compact model architecture. We employ a lightweight model, namely \textit{extensible embedder}, to transform the input into output embeddings. Then we adopt another \textit{down-scaling function}, which down samples the output embeddings by a factor of $k$ (e.g., $k=32$). In other words, one out of $k$ output embeddings are sampled as an extensible embedding. Notably, the down-scaling can be conducted with an arbitrary scaling factor and sampling scheme at the inference time. Thus, it contributes to a high flexibility of practical usage, which enables the ad-hoc extension of different context lengths. 

The extensible embedder is learned through the \textit{two-stream auto-regression (AR)} tasks, where each training sample is processed by two passes of feed-forward. In the first pass, the extensible embeddings are generated and cached for each training sample. In the second pass, the next tokens are predicted based on the extensible embeddings of their preceding contexts. Based on such a tailored for of auto-regression, comprehensive training losses can be derived from all tokens within each training sample. Therefore, it brings forth two benefits: on one hand, the extensible embeddings can be learned to assist the LLM's generation directly, which is well aligned with the downstream LLM's working process; on the other hand, it results in an exceptional sample efficiency, which enables the model to be effectively trained with a small amount of data. 

The training process is performed with the downstream LLM's parameters fixed all the time. Thus, the extensible embeddings can work as a \textit{plug-in module}, which brings extended contextual information without compromising the LLM's original performance with short contexts. Interestingly, we also observe the strong but unexpected compatibility of extensible embedding beyond its downstream LLM. In particularly, the well-trained extensible embeddings for one LLM can be effectively applied to other fine-tuned derivatives of its downstream LLM without any further adaptation. Such a property suggests the extensible embedding's potential as a versatile method for the context extension across a family of closely related LLMs.  

We initialize the extensible embedder with the first 8 layers of LLaMA-2-7B model \cite{touvron2023llama-b}, where it is trained to extend the context for another downstream LLaMA-2-7B model. With just 100K training samples from RedPajama \cite{together2023redpajama} and LongAlpaca \cite{chen2023longlora}, the extensible embedder is able to achieve a superior capability in context extension. Notably, it enables the extension of LLaMA-2-7B (4K) over \underline{\textit{100K}}, while producing superior performances on both long-context language modeling and understanding tasks. Besides, by applying to other fine-tuned derivatives of LLaMA-2-7B with larger context windows, e.g., LongChat-32K \cite{longchat2023}, it can further enable high-quality generation with super-long contexts over \underline{\textit{1 million tokens}}.

To summarize, this work is highlighted for the following contributions. 1) Extensible embedding presents a simple but effective method, which establishes a long context for the LLM based on compact representation of the input. 2) The tailored model architecture facilitates superior and flexible extension for different context lengths, and 3) the sample-efficient two-stream AR task enables the cost-effective training of the model. 4) Comprehensive experiments verify extensible embedding as an \textit{effective}, \textit{efficient}, \textit{flexible}, and \textit{compatible} method for the extension of LLM's context.


\begin{figure*}[t]
    \centering
    \includegraphics[width=0.95\textwidth]{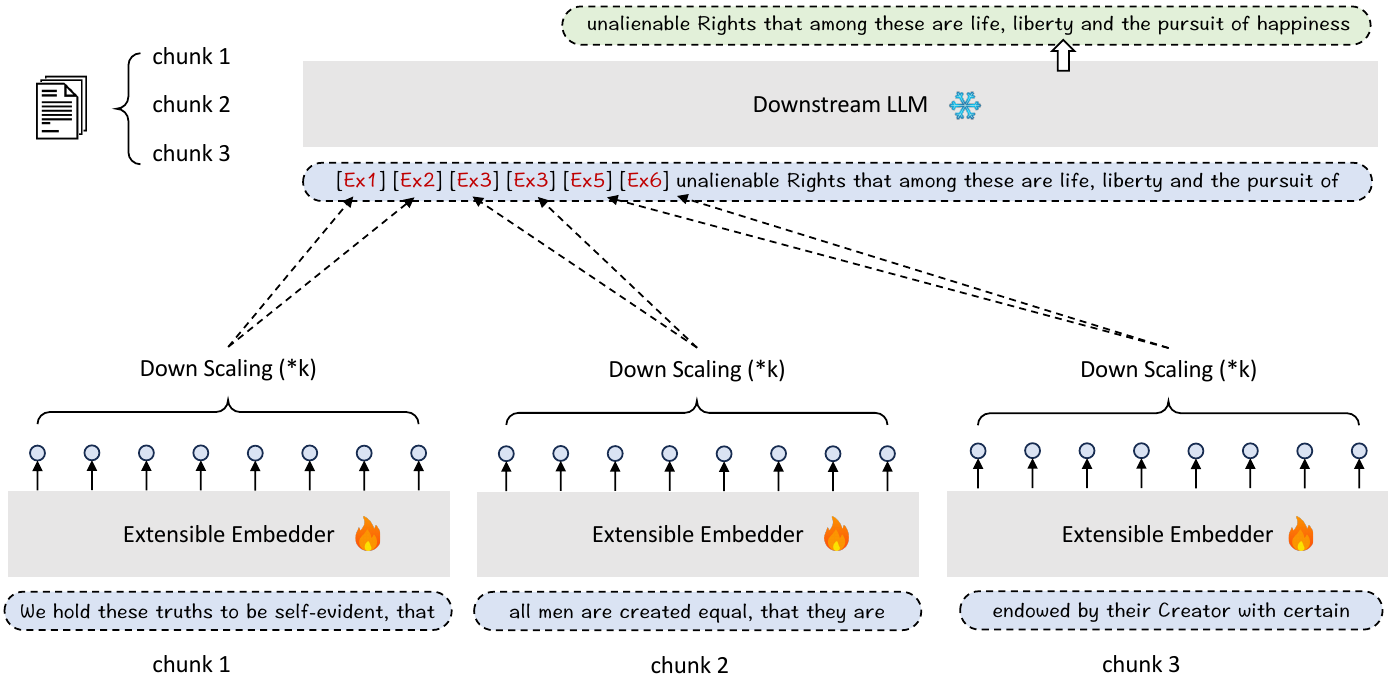}
    \vspace{-5pt}
    \caption{\textbf{Framework}. The input data is partitioned into chunks. Each sub-sequence is transformed and down-scaled as extensible embeddings. The new tokens are predicted based on the extensible embeddings from preceding chunks and the token embeddings in the same chunk. The extensible embedder is learned with a fixed downstream LLM.} 
    \vspace{-10pt}
    \label{fig:supertoken}
\end{figure*} 

\section{Extensible Embedding} 

\subsection{Framework}
The workflow of extensible embedding is shown as Figure \ref{fig:supertoken}, where a long-sequence input $X$ (e.g., a long document of 16K tokens) can be utilized by a LLM (e.g., LLaMA-2) with a short context window (4K). Firstly, the input $X$ is partitioned into chunks: $\{X_{1}, ... X_{N}\}$.  The chunk length $L_i$ is set as the maximum window size of the extensible embedder, e.g., $L_i = 4096$ with LLaMA-2, where the coherence of context can be best preserved. Secondly, each chunk is transformed by the extensible embedder into its output embeddings. The output embeddings are down-scaled by the scaling factor $k$ (e.g., $k=32$), where $L/k$ extensible embeddings (denoted as EX) are produced as the compact representation of the input. Finally, the new tokens are predicted conditioned on the extensible embeddings from the preceding chunks and the normal token embeddings within the recent context. 

\subsection{Embedding Generation}\label{sec:embed}

The typical token embedding, which is corresponding to each individual token, is information sparse. In contrast, the extensible embedding is used to represent an extensible scope of context, e.g., multiple words or sentences, which possess a higher information density. We employ a language model as the embedder ($\mathrm{LM}_{ex}$), which transforms the input $X_i: \{x_{i,1}, ... x_{i,L}\}$ into output embeddings $O_i$: 
\begin{equation*}
    O_i: \{o_{i,1}, ... o_{i,L}\} \leftarrow \mathrm{LM}_{ex}(x_{i,1}, ... x_{i,L}; \theta_{ex}).
\end{equation*}
On top of an expressive embedder, each output embedding can serve as a high-quality representation for its preceding context, i.e. $o_{i,j}$ for $x_{i,1}, ... x_{i,j}$. To acquire the compact representation for the entire input, the output embeddings are further down-scaled by the scaling factor $k$, where $m$ ($m = L/k$) extensible embeddings ($ex_{i,*}$) are generated for $X_i$: 
\begin{equation*}
    \{ ex_{i,1}, ..., ex_{i,m} \} \leftarrow \mathrm{DownScale}(\{o_{i,1}, ... o_{i,L}\}). 
\end{equation*}
There can be many alternative ways to realize the functionality of down-scaling, where arbitrary pooling functions along the sequence dimension can be applied. In our work, we simply down-scale the output embeddings through the strided sampling, where the last embedding in every $k$ steps is chosen, i.e., $ex_{i,j} \leftarrow o_{i,k{\times}j}$. On one hand, such a simple scheme is easy to realize on top of the existing LLM's architecture, and it produces the optimal empirical performance in the downstream tasks. On the other hand, it leads to a high flexibility of usage, where the context can be extended by an arbitrary scaling factor by simply adjusting the downs-sampling rate ($k$) at the inference time.

\begin{figure*}[t]
    \centering
    \includegraphics[width=\textwidth]{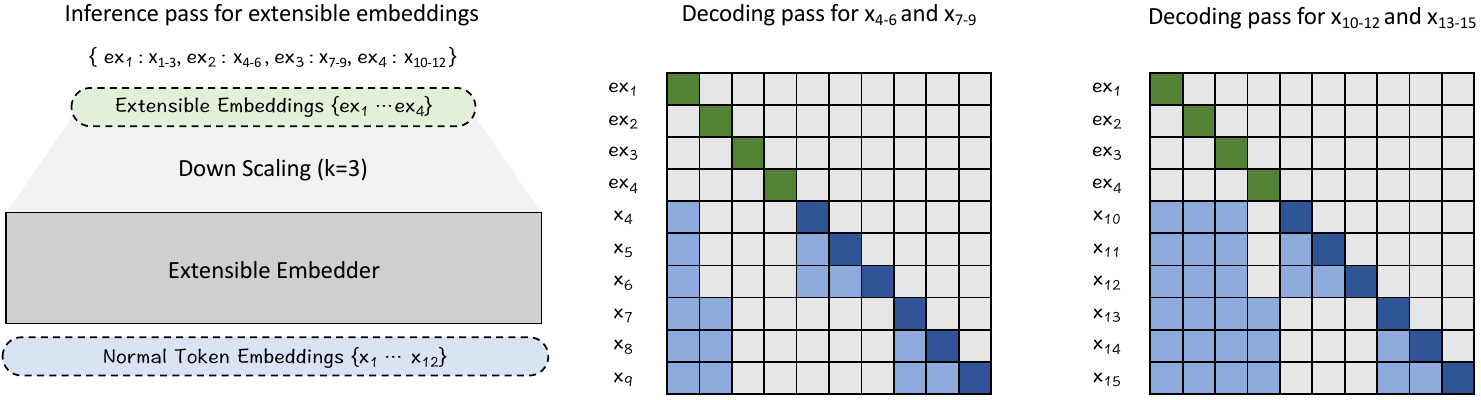}
    \vspace{-15pt}
    \caption{Two-Stream AR. In the first pass, the normal embeddings are transformed into extensible embeddings (with a scaling factor $k=3$). In the second pass (window size 10, chunk size 3), the auto-regression is accomplished in two sliding steps: the $x_{1-3}$ and $x_{4-6}$ predicted in the first step, $x_{7-9}$ and $x_{10-12}$ predicted in the second step.} 
    \label{fig:supertoken_attention}
    \vspace{-10pt}
\end{figure*}

\subsection{Learning Method}\label{sec:learn}
The extensible embeddings are learned by the auto-regression (AR) tasks, where the loss is minimized for the prediction of next tokens conditioned on the extensible embeddings from the preceding context. The auto-regression can be simply performed by having the long context transformed into extensible embeddings and predicting the last few tokens within one training sample, e.g., predicting the answer to a question based on the extensible embeddings of a long document. However, the naive method will be limited by its inferior training effect, because the long context accounts for the majority of computation cost whereas no prediction loss can be produced from it.   

In our work, we propose \textbf{two-stream AR} which trains the model with optimized sample efficiency (Figure \ref{fig:supertoken_attention}). In the first pass of inference, the extensible embeddings are generated for the entire context. For example, with a chunk size of 3 and an scaling factor of 3, the input data $X=\{x_1, ... x_{15}\}$ is transformed into the extensible embeddings $\{ex_{1,1}, ex_{2,1}, ex_{3,1}, ex_{4,1}\}$ (the last chunk is exempted). In the second past, each single token within the long context is streamingly predicted by chunks. Particularly, the prediction is made conditioned on the extensible embeddings from the previous chunks and the preceding normal token embeddings within the same chunk. Formally, 
\begin{align*}
    \min\limits_{\theta_{ex}} \sum\limits_{X}  &  \sum\limits_{i>1} \log P(x_{i,j} |  \\
    & ex_{1,1}, ... ex_{i-1,k}, x_{i,1}, ... x_{i,j-1} ; \theta, \theta_{ex}).
\end{align*}
For example, $x_6$ is predicted based on $ex_{1}$ (representing $x_{1-3}$) and $x_4$. Crucially, the chunk size of training is made much smaller than the LLM's window size (e.g., 512), where the prediction of new tokens can mostly rely on the contextual information offered by the extensible embeddings. Thanks to the above processing, the prediction loss can be comprehensively derived from the each training sample, which enables the model to be effectively learned from a small amount of data. We also randomly sample the extension ratio $k$ from a candidate scope (e.g., [2, 4, 8, 16, 32]) for each training sample, which benefits the model's generalization for the extension of diverse context lengths. 

The extensible embeddings are learned with the downstream LLM's parameters ($\theta$) fixed all the time. As a result, the LLM's original capabilities on short contexts are not affected by the introduction of extensible embeddings. Besides, we also empirically observe the strong but unexpected compatibility from the above training process, where the extensible embeddings can be directly applied to the fine-tuned derivatives of the downstream LLM without further adaptation.

\subsection{Inference}\label{sec:inference}
The inference with the extensible embeddings is discussed w.r.t. the online and offline scenario, respectively. In particularly, the online scenario deals with the situation where the long-sequence data is streamingly presented (e.g., conversation). In this scenario, the generation process is conducted in consecutive sessions. In each session ($i$-th), the LLM predicts the new token ($t_{i,j}$) based on the extensible embeddings from the previous sessions ($Ex_{<{i}}$) and the preceding normal token embeddings within the current session ($\{x_{i,<j}\}$). The current session comes to its end when the total sum of both types of embeddings reaches the maximum capacity of context window ($L^*$): $|Ex_{<{i}}| + j = L^*$. Then, the normal token embeddings $\{x_{i,*}\}$ will be transformed into the extensible embeddings of the current session $Ex_{{i}}$ and used by the next session.   


The offline scenario handles the cases where the long-sequence data is fully presented beforehand (e.g., RAG, reading comprehension of long-docment). In this scenario, the extensible embeddings can be pre-computed for the data, which will significantly benefit the efficiency of online inference. In fact, it is OK to simply save the whole output embeddings from the extensible embedder, and flexibly sample for the extensible embeddings during inference based on the ad-hoc scaling factor. 

\renewcommand{\arraystretch}{1.2}
\begin{table*}[t]
    \centering
    \scriptsize
    \begin{tabular}{l|ccccc|ccccc}
    \ChangeRT{1pt}
        \multirow{2}{*}{\textbf{Model}} & \multicolumn{5}{c|}{PG19} & \multicolumn{5}{c}{Books3} \\
        \cmidrule(lr){2-6} \cmidrule(lr){7-11}
        & \textbf{4K} & \textbf{8K} & \textbf{16K} & \textbf{32K} & \textbf{100K} & \textbf{4K} & \textbf{8K} & \textbf{16K} & \textbf{32K} & \textbf{100K} \\
    \midrule
        LLaMA-2-7B & 7.77 & \textgreater$10^3$ & \textgreater$10^3$ & \textgreater$10^3$ & OOM & 4.21 & \textgreater$10^3$ & \textgreater$10^3$ & \textgreater$10^3$ & OOM \\
        PI & 7.77 & 8.68 & 18.65 & \textgreater$10^2$ & OOM & 4.21 & 5.99 & 11.4 & 69.8 & OOM \\
        NTK & 7.77 & 8.13 & 10.71 & 55.22 & OOM & 4.21 & 5.10 & 7.71 & 52.3 & OOM \\
        StreamingLLM & 7.98 & 8.01 & 8.00 & 8.00 & 8.00 & 4.32 & 4.34 & 4.33 & 4.33 & 4.34 \\
    \midrule
        LongAlpaca-16K & 8.45 & 8.15 & 8.12 & \textgreater$10^3$ & OOM & 4.93 & 4.67 & 4.64 & \textgreater$10^3$ & OOM \\
        LongChat-32K & 7.59 & 7.25 & 7.00 & 6.85 & OOM & 4.12 & 3.95 & 3.87 & 3.85 & OOM \\
        AutoCompressor-6K & 26.9 & \textgreater$10^3$ & $10^3$ & \textgreater$10^4$ & OOM & 17.1 & \textgreater$10^3$ & \textgreater$10^3$ & \textgreater$10^4$ & OOM \\
        LongLLaMA & 7.12 & 6.95 & 6.78 & OOM & OOM & 3.99 & 3.90 & 3.84 & OOM & OOM \\
    \midrule
        ExtEmbedding ($\times16$) & 7.75 & 7.48 & 7.38 & 7.31 & \textgreater$10^2$ & 4.32 & 4.20 & 4.15 & 4.13 & \textgreater$10^3$ \\
        ExtEmbedding ($\times32$) & 8.61 & 8.15 & 7.87 & 7.69 & 7.54 & 4.67 & 4.48 & 4.36 & 4.28 & 4.25 \\
    \ChangeRT{1pt}
    \end{tabular}
    \vspace{-5pt}
    \caption{Language modeling performance (measured by perplexity) on PG19 and Books3.}
    \vspace{-10pt}
    \label{tab:ppl}
\end{table*}

\begin{figure}[t]
    \centering
    \includegraphics[width=\linewidth]{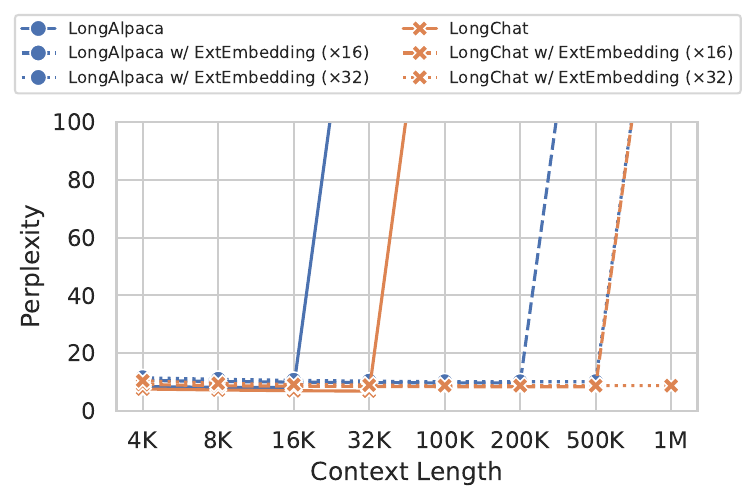}
    \vspace{-20pt}
    \caption{The extensible embedding trained on LLaMA-2-7B can be directly utilized by LongAlpaca-16K and LongChat-32K, leading to the scaling of their context lengths by $\times16$ and $\times32$ (PPL on PG19). Remarkably, the context of LongChat can be extended to 1 million.} 
    \vspace{-15pt}
    \label{fig:ppl_compatibility}
\end{figure}

\section{Experiments}
\vspace{-2pt}
In this section, we conduct comprehensive experiments to investigate the following key issues about extensible embedding. 1) The effectiveness of context extension. 2) The flexibility and compatibility. 3) The running efficiency. 4) The technical factors about extensible embedding. 

\subsection{Experimental Settings}
We leverage LLaMA-2-7B (chat) \cite{touvron2023llama-b} as our downstream LLM. We initialize extensible embedder with the first 8 layers of LLaMA-2-7B (chat). The training takes place on one Nvidia 8×A800 GPU machine, with a batch size of 8 and a learning rate of $5e^{-5}$ using the linear scheduler. The training is consecutively performed with 90K sampled instances from Redpajama \cite{together2023redpajama} and 10K training instances from LongAlpaca \cite{chen2023longlora}. Extensible embedding is trained with the downstream LLM's parameters always fixed. 
We consider the following baselines. The fine-tuning free methods: Positional Interpolation (PI) \cite{chen2023extending}, NTK-Aware Scaled RoPE (NTK) \cite{ntkaware2023}, StreamingLLM \cite{xiao2023efficient}. The fine-tuned full-attention methods: LongAlpaca-7B-16K \cite{chen2023longlora}, LongChat-7B-32K \cite{longchat2023}. The fine-tuned methods with modified architectures for long context: AutoCompressor-7B-6K \cite{raecompressive2019}, LongLLaMA-7B \cite{tworkowski2023focused}. All baselines are based on LLaMA-2-7B, except LongLLaMA which leverages CodeLLaMA \cite{roziere2023code}. 


\begin{table*}[t]
    \centering
    \scriptsize
    \begin{tabular}{l|c|cccc|cccc|cccc}
    \ChangeRT{1pt}
        \multirow{2}{*}{\textbf{Model}} & \multirow{2}{*}{\textbf{Len.}} & \multicolumn{4}{c|}{\textbf{Single-Doc QA}} & \multicolumn{4}{c|}{\textbf{Multi-Doc QA}} & \multicolumn{4}{c}{\textbf{Summarization}} \\
        \cmidrule(lr){3-6} \cmidrule(lr){7-10} \cmidrule(lr){11-14}
        & & \textbf{NQA} & \textbf{QASP} & \textbf{MF} & \textbf{Avg.} & \textbf{HQA} & \textbf{2WIKI} & \textbf{MSQ} & \textbf{Avg.} & \textbf{GOV} & \textbf{QS} & \textbf{MN} & \textbf{Avg.} \\
    \midrule
        LLaMA-2-7B & 4K & 18.70 & 19.20 & 36.80 & 24.90 & 25.40 & 32.80 & 9.40 & 22.60 & 27.30 & 20.80 & 25.80 & 24.70 \\
        PI & 16K & 12.85 & 20.86 & 23.24 & 18.98 & 22.41 & 22.13 & 6.94 & 17.16 & 29.77 & 18.48 & 26.85 & 25.03 \\
        NTK & 16K & 15.96 & 19.26 & 34.41 & 23.21 & 29.16 & 29.06 & 11.79 & 23.34 & 29.56 & 17.44 & 26.19 & 24.40 \\
        StreamingLLM & 4K & 18.95 & 16.84 & 26.16 & 21.47 & 30.09 & 26.87 & 11.03 & 22.22 & 25.81 & 20.64 & 25.88 & 22.20 \\
    \midrule
        LM. w. ExtEmbedding* & 4K & 19.59 & 23.18 & 33.90 & 25.56 & 34.53 & 32.93 & 13.30 & 26.92 & 27.06 & 21.15 & 25.69 & 24.63 \\
    \midrule
    \midrule
        LongAlpaca-16K (4k) & 4K & 17.85 & 27.66 & 34.91 & 26.81 & 29.50 & 31.32 & 12.50 & 24.44 & 30.09 & 23.12 & 27.57 & 26.93 \\
        LongAlpaca-16K (8k) & 8K & 18.60 & 28.89 & 38.35 & 28.61 & 32.32 & 30.84 & 11.34 & 24.83 & 31.43 & 24.42 & 27.87 & 27.91 \\
        LongAlpaca-16K & 16K & 19.13 & 28.91 & 37.03 & 28.36 & 36.93 & 30.32 & 17.23 & 28.16 & 31.30 & 24.16 & 27.84 & 27.77 \\
    \midrule
        LA. w. ExtEmbedding* & 4K & 20.12 & 29.45 & 36.25 & 28.61 & 37.58 & 33.54 & 13.84 & 28.32 & 30.45 & 22.66 & 27.52 & 26.88 \\
    \midrule
    \midrule
        LongChat-32K (4k) & 4K & 15.30 & 27.81 & 41.30 & 28.14 & 28.33 & 25.00 & 12.30 & 21.88 & 31.67 & 21.66 & 26.44 & 26.59 \\
        LongChat-32K (8k) & 8K & 17.35 & 29.14 & 41.67 & 29.39 & 29.58 & 24.65 & 10.83 & 21.69 & 32.40 & 22.30 & 26.39 & 27.03 \\
        LongChat-32K (16k) & 16K & 20.82 & 29.19 & 42.53 & 30.85 & 31.78 & 25.04 & 13.16 & 23.33 & 31.28 & 22.65 & 26.44 & 26.79 \\
        LongChat-32K & 32K & 21.00 & 29.25 & 42.70 & 30.98 & 32.99 & 24.86 & 14.02 & 23.96 & 31.03 & 23.00 & 26.44 & 26.82 \\
    \midrule
        LC. w. ExtEmbedding* & 4K & 17.07 & 30.59 & 42.69 & 30.12 & 31.52 & 25.84 & 13.17 & 23.51 & 31.19 & 20.62 & 26.77 & 26.19 \\
    \ChangeRT{1pt}
    \end{tabular}
    \vspace{-5pt}
    \caption{The evaluation of long-context understanding with tasks from LongBench. ``(*k)'' indicates that the input data is truncated to *k for the corresponding model. (LM.: LLaMA-2-7B, LA.: LongAlpaca, LC.: LongChat)} 
    \vspace{-15pt}
    \label{tab:longbench}
\end{table*}

\subsection{Language Modeling}
The long-context language modeling is evaluated with PG19 \cite{raecompressive2019} and Books3 \cite{pile}. Following the method used by Alexis et al. \cite{chevalier2023adapting}, 
the perplexity is measured by predicting the last 512 tokens based on the preceding context. There are two evaluation settings about the extensible embedding: ExtEmbedding ($\times16$) and ExtEmbedding ($\times32$), where the scaling factor $k$ is set as 16 and 32, respectively. The evaluation results are shown in Table \ref{tab:ppl}, where the following observations can be derived. 

On the one hand, we can observe the superior long-context language modeling quality achieved by extensible embedding. Firstly, extensible embedding leads to a notable improvement over the LLaMA-2-7B baseline, which indicates that the extended context can be effectively utilized to improve the generation quality. Secondly, the relative improvement (over LLaMA-2-7B) from extensible embedding is more pronounced than the fine-tuning free method. Although the fine-tuned full-attention methods may produce better performances in some cases, they require the change of the LLM's original parameters, and work with much higher running costs. Thirdly, extensible embedding is able to flexibly support much longer contexts by simply adjusting the scaling factor ($k$). In particular, by increasing the scaling factor from $16$ to $32$, LLaMA-2-7B's context length can be continually extended beyond $100K$ (up to $32 \times 4K$). The above observations validate the {effectiveness of context extension} with extensible embedding. 

On the other hand, we can also make interesting observations about the extensible embedding's compatibility with the fine-tuned derivatives of its downstream LLM. We utilize two baseline models for evaluation: LongAlpaca and LongChat. Both models are fine-tuned from LLaMA-2-7B with long-sequence data, which achieve longer context windows of 16$K$ and 32$K$, respectively. As we can observe from Figure \ref{fig:ppl_compatibility}, the well-trained extensible embeddings on LLaMA-2-7B can be directly applied to the two models without any adaptation, which scales up their contexts by 16$\times$ and 32$\times$ times (with ExtEmbedding 16$\times$ and 32$\times$, respectively). Remarkably, we can reach a context length of 1 million tokens by enhancing LongChat-32$K$ with ExtEmbedding (32$\times$). We make further exploration with more fine-tuned derivatives (in Appendix \ref{app:compatibility}) and different evaluation tasks (\S \ref{sec:exp-long-under}), whose results affirm the ubiquity of this property.

\subsection{Language Understanding}\label{sec:exp-long-under}
We evaluate long-context language understanding using 9 datasets from LongBench \cite{bai2023longbench}, which are about single-doc QA, multi-doc QA, and summarization. 
It's worth noting that the sequence lengths for majority of the evaluation samples are less than 16K or 32K. Therefore, the performance from the two fine-tuned methods LongAlpaca and LongChat can almost be a upper-bound for the rest of the methods. For each evaluation sample, the scaling factor is adjusted case-by-case for extensible embedding, which will let the compressed input just fit into the 4K context window of LLaMA-2-7B. The evaluation results are presented in Table \ref{tab:longbench}, where the following observations can be made. 

\begin{table*}[t]
    \centering
    \scriptsize
    \begin{tabular}{l|ccccc|ccccc}
    \ChangeRT{1pt}
        \multirow{2}{*}{\textbf{Model}} & \multicolumn{5}{c|}{\textbf{GPU Memory (GB)}} & \multicolumn{5}{c}{\textbf{Inference Time (s)}} \\
        \cmidrule(lr){2-6} \cmidrule(lr){7-11} 
        & 4K & 8K & 16K & 32K & 100K & 4K & 8K & 16K & 32K & 100K \\
    \midrule
        LongChat-32K & 18.12 & 23.68 & 34.79 & 57.03 & OOM & 0.32 & 0.65 & 1.43 & 3.32 & OOM \\
        StreamingLLM
        & 15.11 & 15.11 & 15.11 & 15.11 & 15.11 & - & - & - & - & - \\
        LongLLaMA & 17.73 & 21.40 & 33.41 & OOM & OOM & 0.60 & 1.44 & 3.30 & OOM & OOM \\
    \midrule
        ExtEmbedding (online) & 20.33 &21.59 & 21.59 & 21.59 & 21.59 & 0.28 & 0.49 & 0.86 & 1.57 & 3.43 \\
        ExtEmbedding (offline) & 13.96 & 14.21 & 14.75 & 15.79 & 17.54 & 0.08 & 0.08 & 0.10 & 0.12 & 0.23 \\
    \ChangeRT{1pt}
    \end{tabular}
    \vspace{-5pt}
    \caption{Efficiency analysis in terms of GPU memory usage and inference time.}
    \vspace{-15pt}
    \label{tab:memory&time}
\end{table*}



Firstly, extensible embedding can substantially improve upon the LLaMA-2-7B baseline for both single-doc QA and multi-doc QA. By comparison, the fine-tuning free methods bring in almost no contribution or even negative effect to long-context understanding, despite their effectiveness in language modeling. Although the fine-tuned methods bring seemingly better results on both QA and summarization tasks, we find that the improvement is mainly from the LLM's improved performance with short contexts (due to fine-tuning) rather than the incorporation of longer contexts. Particularly, both LongAlpaca and LongChat already achieve sufficiently high performances with the basic 4k context. For summarization, the further extension of context length is of little benefit. To some extent, the summarization cannot properly reflect the long context capability as most of the useful information for summarization is presented in the beginning or the end of each document, which has been covered by the basic 4K context. In the sense of relative improvement purely from the extended context, the extensible embedding's effect is comparable with fine-tuning. Meanwhile, it preserves a higher efficient because it only takes a 4K context. 


Secondly, extensible embedding remains compatible with the fine-tuned derivatives of LLaMA-2-7B in this scenario. We directly apply the well-trained extensible embedding for LongAlpaca and LongChat. As introduced, the sequence lengths for the majority of evaluation samples are less than 16K or 32K. For those cases, the extensible embeddings will not introduce extra context, but only compress their original context to 4K, which makes the inference process more efficient. Notably, the two model's strong performances with the full-scale contexts can be effectively preserved by LA./LC. w. ExtEmbedding, which indicates two interesting properties: 1) the extensible embedding presents an almost lossless compression of the context, 2) the well-trained extensible embedding for LLaMA-2-7B can be seamlessly transferred to LongAlpaca and LongChat. 

\vspace{-2pt}
\subsection{Efficiency Analysis}
We analyze the efficiency in terms of GPU memory usage and inference time. The experiment is on a single Nvidia A800-80G GPU. The performance is measured by taking the average value of 100 forward passes where the last 512 tokens are predicted based on the input context. 
We include the following baselines. LongChat based on full-attention, where FlashAttention-2~\cite{dao2023flash} is enabled for its acceleration. StreamingLLM based on stream processing, whose window size is set to 2048; it is exempted from time evaluation because its current stepwise implementation is too slow. 
The extensible embedding uses a scaling factor $k=32$, where both working modes are evaluated: the online mode where data is streamingly presented and extensible embeddings are generated step-by-step; the offline mode where data is presented in advance and extensible embeddings are pre-computed. The following observations can be made from Table \ref{tab:memory&time}. 

First of all, the online mode leads to a constant memory usage, while the offline mode results in an even smaller consumption. As introduced (\S \ref{sec:inference}), the memory usage of extensible embedding comes from two parts. The first part is the generation of extensible embeddings, where the stream processing is conducted with a 4K sliding window (the offline mode is free from this step, thus taking even less GPU memory). The second part is the final inference stage based on the extensible embeddings, where the input sequence has been substantially condensed and become much shorter than its original length. Because the two operations are consecutively conducted, their memory costs will not accumulate. The online mode deals with both parts, but the overall memory cost is dominated by the first part. The offline mode only needs to handle the second part, which results in an even smaller cost. Both modes are free from processing the entire long input simultaneously, which contributes to a very economic usage of GPU memory.

Secondly, extensible embedding exhibits a much smaller time cost compared with the baseline methods. 
For ExtEmbedding (online), the majority of its computation is spent on the generation of extensible embedding. Because of the stream processing, the growth of its inference time is linear to the sequence length. Besides, with the pre-computation of the extensible embeddings, the inference time can be dramatically reduced for the offline mode. Such a superior time efficiency will substantially benefit extensible embedding's application in scenarios like RAG and long-doc QA, where long-sequence data can be presented in advance.

\subsection{Ablation Studies}
We perform ablation studies to analysze the influential factors about extensible embedding, whose results are presented with Table \ref{tab:ablation}, \ref{tab:cost_effectiveness}, and \ref{tab:ppl_flexibility}. 


Firstly, we explore the impact of down-scaling by comparing the default strided down-sampling (\S \ref{sec:embed}) with: 1) random down-sampling, which randomly chooses $L/k$ output embeddings from the embedder, 2) terminal down-sampling, which selects the last $L/k$ output embeddings from the embedder ($L$: chunk size). On one hand, the default strided down-sampling method outperforms the two baselines probably due to its more effective coverage of the context. On the other hand, the baselines can be directly applied to the embedder trained with strided down-sampling, which reflects the flexibility of usage of extensible embedding. 

Secondly, we analyze the impact from the size of embedder. Our default method uses the first 8 layers of LLaMA-2-7B, while the baseline uses the first 4 layers. It can be observed that the improved size leads to a better performance. There is no surprise about this observation because a larger embedder is more expressive and able to produce a better representation of the context. Nevertheless, it also comes with a larger computation cost. The optimal trade-off between cost and effectiveness must be determined for each scenario case-by-case. 

\begin{table}[t]
    \centering
    \scriptsize
    \begin{tabular}{l|l|c|c}
    \ChangeRT{1pt}
        \textbf{Factor} & \textbf{Setting} & \textbf{PG19} & \textbf{QA} \\
    \midrule
        \multirow{3}{1.5cm}{Down scaling} & Random down-sampling & 7.64 & 23.39 \\
        & Terminal down-sampling & 7.58 & 24.04 \\
        & Strided down-sampling* & 7.31 & 25.56 \\ 
    \midrule
        \multirow{2}{1.5cm}{Embedder size} 
        & First 4-layer (Llama-2-7B) & 7.46 & 23.32 \\
        & First 8-layer (Llama-2-7B)* & 7.31 & 25.56 \\
    \midrule
        \multirow{2}{1.5cm}{Scale sampling} 
        & Monotonous ($k = 16$) & 7.29 & 21.37 \\
        & Dynamic Sampling* & 7.31 & 25.56 \\        
    \ChangeRT{1pt}
    \end{tabular}
    \vspace{-5pt}
    \caption{Ablation studies. PG19 is measured by PPL under a 32K context; Single-Doc QA is measured by F1 score. Default settings are marked with ``*''. }  
    \vspace{-15pt}
    \label{tab:ablation}
\end{table}

Thirdly, we study the necessity of dynamically sampling the scaling factor during training (Scale sampling, \S \ref{sec:learn}). As a comparison, we employ a constant scaling factor $k$=$16$ (Monotonous). The Monotonous baseline achieves a comparable PPL on PG19 as our default method, because the language modeling task only performs a constant scaling down of the context by a factor of $k$=$16$. However, the default method notably outperforms Monotonous on Single-Doc QA, which relies on diversified scaling factors to condense the input of different lengths for a 4K context window. 

We further investigate the impact of our training method based on two-stream AR (\S \ref{sec:learn}), where two common strategies are introduced as our baselines \cite{mu2023learning,chevalier2023autocompressors,ge2023context}: 1) auto-encoding, where the input data is encoded in the first place and then decoded from the encoding result; 2) text continuation, where the head of input data (the first half) is encoded and the remaining part of the data is decoded from the encoding result. The experiment results are shown in Table \ref{tab:cost_effectiveness}, where two-stream AR notably outperforms the baselines. In just 100 steps, two-stream AR is able to achieve a remarkable performance of language modeling on PG19, which verifies it superior sample efficiency of training. 

Finally, we make ad-hoc selection of scaling factor ($k$) and benchmark the PPL on PG19 at different context lengths (Table \ref{tab:ppl_flexibility}). On one hand, a smaller scaling factor, which means less compression of the data, can preserve a better generation quality. On the other hand, a larger scaling factor, which means higher compression of the data, will achieve a longer extension of the context. Consequently, the scaling factor should be properly selected in practice, such that the needed context can be fully covered with the lowest level of data compression.

\section{Related Works}
\vspace{-2pt}
The extension of LLM's context is a critical issue. Recently, numerous methods have been proposed to tackle this problem from different perspectives. The popular approaches involve the modification of position encoding, e.g., Position Interpolation \cite{chen2023extending} and NTK-Aware \cite{ntkaware2023}, which allows the LLMs to work with new positions during the inference time. The context extension quality from the modified position encoding can be improved by fine-tuning over long-sequence data \cite{peng2023yarn}. However, the fine-tuning is expensive, even with accelerations like LoRA \cite{chen2023longlora,hu2021lora} and sparse attention \cite{chen2023longlora,child2019sparse_transformers}. Besides, the fine-tuning operation may also bring unfavorable effect to the LLM's existing capability. 

\begin{table}[t]
    \centering
    \scriptsize
    \begin{tabular}{l|ccc} 
    \ChangeRT{1pt}
        \multirow{2}{*}{Training method} & \multicolumn{3}{c}{PPL (PG19) at different steps}
        \\ 
        \cmidrule(lr){2-4}
        & {100} & {500} & {1,000} \\
        
    \midrule
        Text Continuation & 32.41 & 11.96 & 11.89 \\
        Auto-Encoding & 10.27 & 9.95 & 10.25 \\
        Two-Stream AR* & 8.85 & 8.17 & 8.00  \\
    \ChangeRT{1pt}
    \end{tabular}
    \vspace{-5pt}
    \caption{Impact from different training methods.}
    \vspace{-10pt}
    \label{tab:cost_effectiveness}
\end{table}

\begin{table}[t]
    \centering
    \scriptsize
    \begin{tabular}{l|ccccc}
    \ChangeRT{1pt}
        \multirow{2}{*}{Scaling factor} & \multicolumn{5}{c}{Context Length} \\
        \cmidrule(lr){2-6}
        & {4K} & {8K} & {16K} & {32K} & {100K} \\
    \midrule
        $k=2$ & 6.80 & 11.76 & \textgreater$10^2$ & \textgreater$10^2$ & \textgreater$10^2$ \\
        $k=4$ & 7.05 & 6.89 & 23.35 & \textgreater$10^2$ & \textgreater$10^2$ \\
        $k=8$ & 7.40 & 7.18 & 7.13 & 27.10 & \textgreater$10^2$\\
        $k=16$ & 7.75 & 7.48 & 7.38 & 7.31 & \textgreater$10^2$ \\
        $k=32$ & 8.61 & 8.15 & 7.87 & 7.69 & 7.54 \\
    \ChangeRT{1pt}
    \end{tabular}
    \vspace{-5pt}
    \caption{Impact from different scaling factors.}
    \vspace{-15pt}
    \label{tab:ppl_flexibility}
\end{table}

In addition to the increasing of window size, people also explore different methods to process a long context with a short context window. One common strategy is to leverage sliding windows \cite{chen2023extending,han2023lm_infinite}, where the long context can be streamingly processed. However, the typical stream processing will simply ignore the information beyond the context window instead of making use of it. Another line of research is about context compression, which follows the same spirit as our method. In general, the context can be compressed in two optional ways. One is to explicitly compress the input data with methods like, summarization \cite{jiang2023llmlingua}, extraction \cite{jiang2023llmlingua}, or retrieval \cite{xu2023retrieval_meets}. Despite simplicity, the explicit compression is prone to incomplete and incoherent contextual information. The other option is to implicitly compress the input into latent embeddings \cite{bulatov2022recurrent,mu2023learning,chevalier2023autocompressors,ge2023context}. The performance of implicit methods highly depend on the quality of compression, which is a joint result from the architecture of compressor and the learning method. So far, none of the previous methods are able to effectively realize a dramatic extension of LLM's context as extensible embedding (longer than 100K) due to the substantial loss of compression. The previous methods also lack the flexibility to support different context lengths. Besides, many of them need modifications on model architectures, which can be incompatible with the existing LLMs.

\section{Conclusion}
In this paper, we present extensible embedding as a new method to extend the LLM's context. It presents a compact representation for an extensible scope of context, which let the LLM to fully perceive the long-sequence input with its limited context window. It is realized based on a flexible model architecture and sample-efficient training algorithm, which not only optimizes the quality of context extension, but also leads to a remarkable flexibility and compatibility of usage, as well as a high efficiency of training and inference. The effectiveness of our method is verified with comprehensive evaluations, where the LLM's context can be dramatically extended with a superior quality.


\clearpage



\bibliographystyle{acl2024}
\bibliography{acl2024} 

\newpage
\appendix
\onecolumn

\section{The Overall Comparison of Extensible Embedding}

\begin{figure*}[htb]
    \centering
    \includegraphics[width=\textwidth]{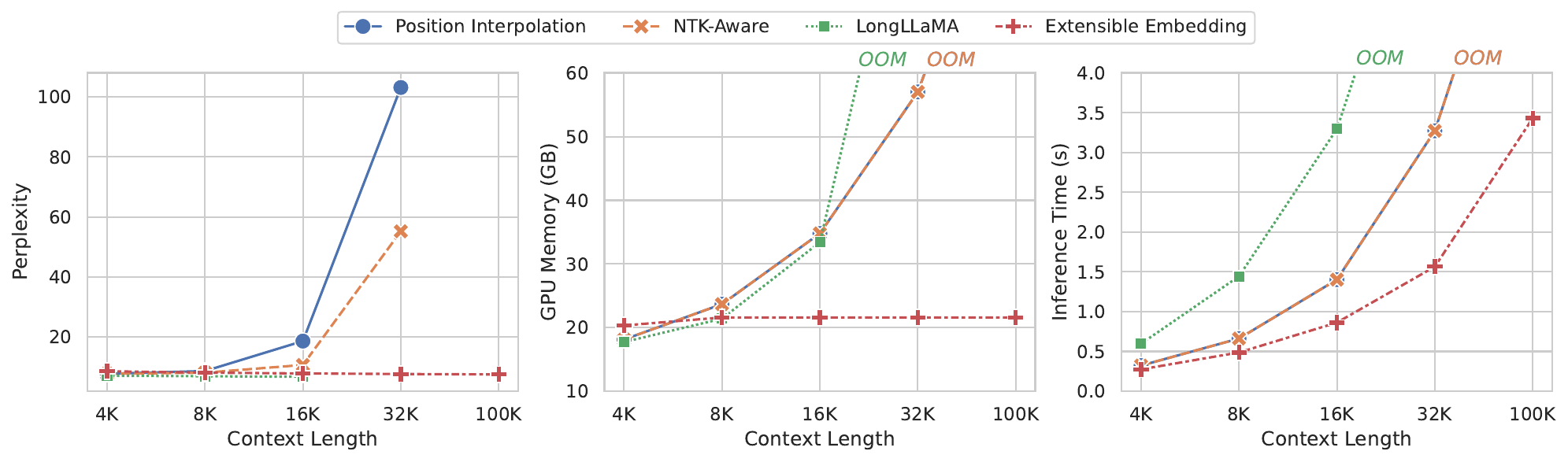}
    \caption{Comparison between extensible embedding and other context extension methods, including 1) Position Interpolation \cite{chen2023extending}, 2) NTK-Aware Scaled RoPE \cite{ntkaware2023}, 3) LongLLaMA \cite{tworkowski2023focused}. 
    Extensible Tokenization presents a superior long-context language modeling capability, along with better efficency in terms of memory and time. 
    Perplexity is measured on PG19 \cite{raecompressive2019} following the method in \cite{chevalier2023adapting}}
    \label{fig:cmp_3}
\end{figure*}

\section{The Compatibility of Extensible Embedding}
\label{app:compatibility}

\begin{figure*}[htb]
    \centering
    \includegraphics[width=\textwidth]{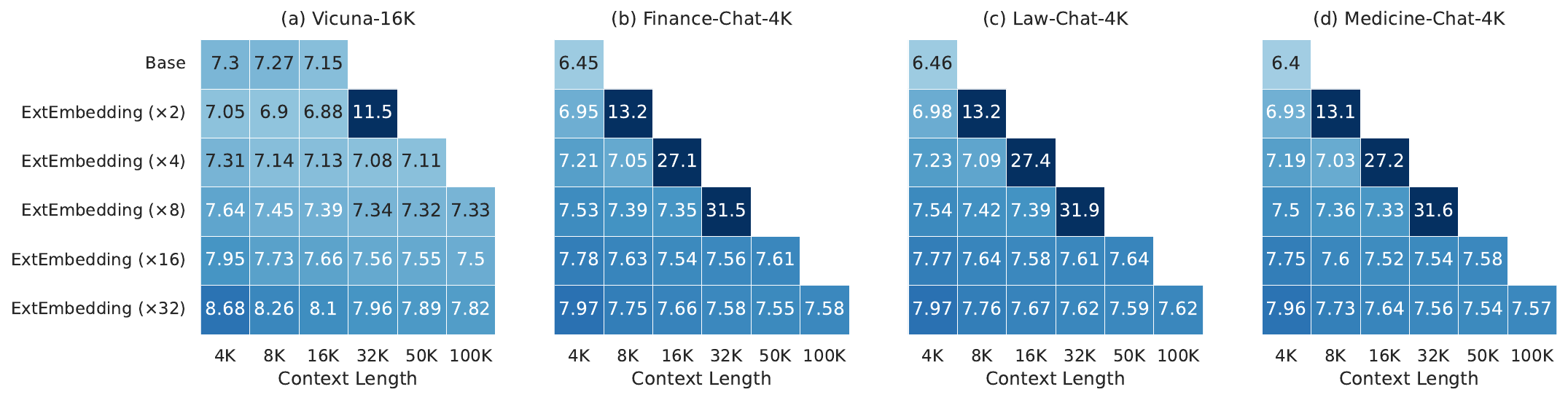}
    \caption{Compatibility of extensible embedding. The metric is perplexity on PG19. The darker the color, the higher the perplexity. And the blank area denotes instance where perplexity $> 10^2$.}
    \label{fig:ppl_compatibility_more}
\end{figure*}

To further investigate the compatibility of extensible embedding, we directly apply the well-trained extensible tokenizer to more LLaMA-2-7B based model (Figure \ref{fig:ppl_compatibility_more}). We select the following models: 1) Vicuna-16K \cite{zheng2023judging}, 2) Finance-Chat-4K \cite{cheng2023adapting}, 3) Law-Chat-4K \cite{cheng2023adapting}, and 4) Medicine-Chat-4K \cite{cheng2023adapting}. All these models are not only popular in the community, but have also been fine-tuned for specific domains. The method for measuring perplexity is consistent with main text. 

Firstly, all these models extend their context length after implementing extensible embedding. For instance, with extensible embedding, Vicuna-16K's context length can extend up to 100K. 
Secondly, as the increase of the context length, the context generation quality improves.
Thirdly, we find all these fine-tuned models perform well under different scaling factors, which indicates that extensible embedding maintains its flexibility when directly applying it to fine-tuned models.

\end{document}